\pdfoutput=1
\documentclass[runningheads]{llncs}

\usepackage[T1]{fontenc}
\usepackage{amsmath}
\usepackage{amssymb}
\usepackage{marvosym}
\usepackage{graphicx}
\usepackage{algorithm}
\usepackage{algpseudocode}
\usepackage{booktabs}
\usepackage[table]{xcolor}
\usepackage[normalem]{ulem}
\usepackage{url}
\usepackage[most]{tcolorbox}
\usepackage{placeins}
\definecolor{caseQ}{HTML}{EEF2F7}
\definecolor{caseBad}{HTML}{FBE7E5}
\definecolor{caseGood}{HTML}{E6F1E8}
\definecolor{caseHeader}{HTML}{E9ECEF}
\definecolor{caseWarn}{HTML}{FFF4D6}
\definecolor{typeGray}{HTML}{EFEFEF}
\definecolor{okGreen}{HTML}{2E7D43}
\definecolor{badRed}{HTML}{C0392B}

\newtcblisting{promptbox}[1]{
  enhanced,
  breakable,
  listing only,
  colback=caseQ,
  colframe=caseHeader,
  coltitle=black,
  title={#1},
  fonttitle=\bfseries\sffamily,
  boxrule=0.5pt,
  arc=1mm,
  left=1mm,
  right=1mm,
  top=1mm,
  bottom=1mm,
  listing options={
    basicstyle=\ttfamily\scriptsize,
    breaklines=true,
    columns=fullflexible,
    keepspaces=true,
    showstringspaces=false
  }
}

\begin{document}

\title{ClueWeaver: Reward-Guided Dual-Agent Evidence Reasoning for Compact LLMs on Literary Long Narratives}
\titlerunning{ClueWeaver}

\author{
\hspace*{-0.055\textwidth}\makebox[1.11\textwidth][c]{Jihao Zhu\inst{2}\thanks{Equal contribution.}, Zhiwei Yang\inst{1}\textsuperscript{*}, Wenxiao Zhang\inst{3}\textsuperscript{*}, Junqian Zhao\inst{1}, Qi You\inst{1}, Fangqi}\hspace*{-0.055\textwidth}\\[-1pt]
\hspace*{-0.055\textwidth}\makebox[1.11\textwidth][c]{Wang\inst{1}, Zheyuan Deng\inst{4}, Hanzhe Yang\inst{3}, Yu Liu\inst{1}\textsuperscript{(\Letter)}, Jin B. Hong\inst{3}\textsuperscript{(\Letter)}}\hspace*{-0.055\textwidth}}
\authorrunning{J. Zhu et al.}
\institute{
Institute of Information Engineering, CAS, Beijing, China\\
\email{liuyu@iie.ac.cn} \and
University of Aberdeen, Aberdeen, United Kingdom \and
The University of Western Australia, Perth, Australia\\
\email{jin.hong@uwa.edu.au} \and
Brown University, Providence, Rhode Island, United States}

\maketitle

\begin{abstract}
Humanities and social science research requires close reading of long
narrative materials such as novels, scripts, archives, and case reports, yet
many users have limited access to costly proprietary long-context models.
Compact, locally deployable language models are a practical alternative, but
directly feeding them an entire long context remains costly, hard to inspect,
and prone to missing sparse evidence. We present \textbf{ClueWeaver}, an evidence-aware
dual-agent framework for long-narrative question answering with compact local
models. A Finder identifies passages containing answer-critical clues through
retrieval-guided segmentation, while an Interpreter derives the answer from the
selected evidence, produces rationales with paragraph-ID citations, and applies
an internal self-calibration pass for high-risk questions.
Both agents are optimized with
reward-guided reinforcement learning: Finder rewards emphasize evidence
retention and faithful paragraph-ID references, and Interpreter rewards
emphasize correctness, grounding, and concise explanations. This decomposition
makes evidence selection and reasoning more inspectable than end-to-end
prompting. Experiments across multiple long-context narrative
question answering and claim verification settings show that ClueWeaver
substantially improves local end-to-end language models while providing evidence
coverage and paragraph-referenced reasoning traces. Code is available at
\url{https://github.com/Ameame1/ClueWeaver}.
\end{abstract}

\section{Introduction}

Novels, screenplays, investigation records, and long case reports require
coherent reading over extended
narratives~\cite{kocisky2018narrativeqa,wang2025novelqa,xu2025detectiveqa,karpinska2024one}.
Rather than collections of independent facts, their meaning emerges from
characters, events, motives, temporal order, and causal links distributed over
hundreds of paragraphs. This setting is especially important for
resource-constrained literary and humanities research, where scholars may need to
analyze novels, scripts, archives, or case materials without costly proprietary
long-context models or remote services~\cite{akazawa2025literary,widder2024open}.
Long-narrative question answering therefore offers a practical testbed for
strengthening compact, locally deployable models for long-context
reading~\cite{yang2025qwen3,mistral2025ministral3,openai2025gptoss}. Recent large
language models, unlike earlier pretrained models with short context windows,
can now accept much longer inputs in a single
prompt~\cite{bai2024longbench,zhang2024inftybench,bai2025longbenchv2}. Yet
window size alone does not ensure reliable narrative
understanding~\cite{liu2024lost,bai2025longbenchv2}: narrative questions often require
sparse, indirect clues spread across distant text
spans~\cite{pang2022quality,xu2025detectiveqa}. Models must therefore identify
salient story evidence.

\begin{figure}[t]
\centering
\includegraphics[width=1\textwidth]{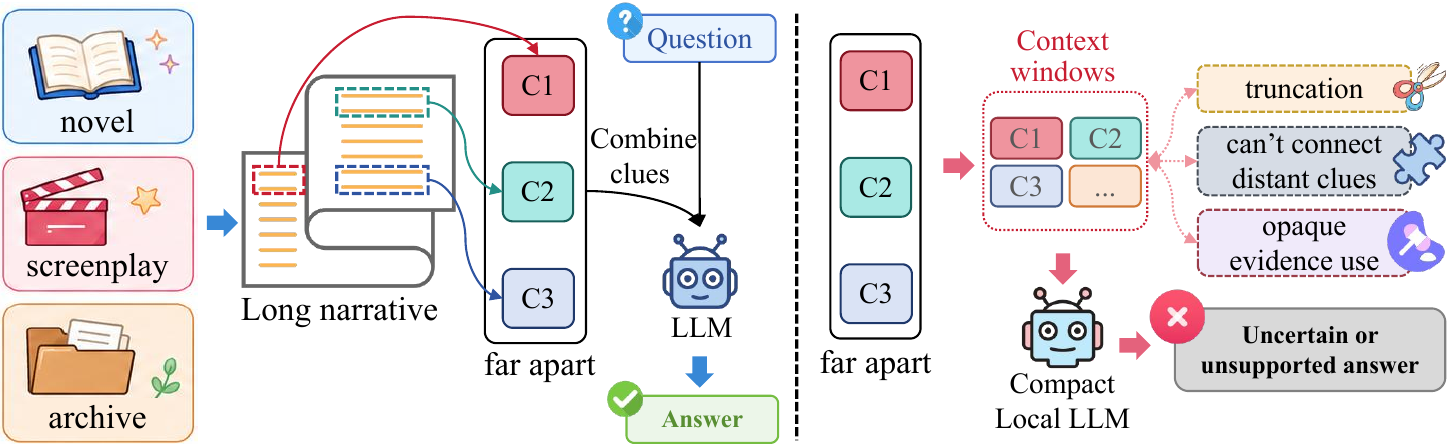}
\caption{Answer-critical clues in long narratives are often
sparse and distant. While long-context models may connect them directly,
compact local models operating over limited context windows can truncate
evidence, separate related clues, and obscure evidence use.}
\label{fig:motivation}
\end{figure}

Figure~\ref{fig:motivation} illustrates this difficulty. In long narratives,
answer-critical clues are often sparse and far apart, requiring models to
connect them in narrative order. Compact local models may instead rely on
limited context windows, which can truncate evidence, separate related clues,
and obscure evidence use. As a result, failures may arise from missing or
misusing key clues, not merely from generation errors.

This setting differs from standard retrieval-augmented generation
(RAG)~\cite{lewis2020retrieval}. While typical RAG retrieves external documents
from a large corpus, long-narrative question answering assumes that the source
text is already given and requires the model to find, preserve, and use sparse
evidence within it. Direct full-context prompting is a natural baseline, but it
keeps evidence selection implicit~\cite{liu2024lost,bai2025longbenchv2}: failures
may result from missing relevant clues, discarding them among irrelevant
details, or failing to reason over available evidence. This raises a central
question: how can a compact local model answer long-narrative questions while
making its supporting clues explicit?

Existing work has made this challenge visible, but does not fully answer the
question above. Long-narrative and long-document benchmarks show that questions
over books, scripts, and passages require more than local
phrase matching or shallow salience~\cite{kocisky2018narrativeqa,pang2022quality}.
Broader long-context evaluations further show that simply increasing input
length does not ensure robust use of relevant information, especially when
evidence is buried inside the context~\cite{bai2024longbench,liu2024lost}.
RAG improves knowledge-intensive generation by
retrieving external passages~\cite{lewis2020retrieval}. IRCoT interleaves
retrieval with chain-of-thought reasoning~\cite{trivedi2023interleaving},
Self-Ask decomposes questions into follow-up queries~\cite{press2023measuring},
Chain-of-Agents distributes long-context reading across collaborating
agents~\cite{zhang2024chain}, and RAG-DDR optimizes RAG modules with
data-driven rewards~\cite{li2025rag}. However, these methods mainly target
open-domain retrieval, general multi-hop question answering, or broad
long-context aggregation. They do not directly address the setting where the
source narrative is already given, answer-critical clues are sparse and
distributed, and evidence selection itself should be explicit and inspectable.
Full-context prompting and single-stage readers, meanwhile, leave evidence selection
implicit and couple evidence locating with answer generation. As a result, they
provide limited control over whether the model is reasoning from the right
evidence. A more suitable framework should expose evidence selection and
evidence-grounded reasoning as separate, inspectable, and trainable steps.

To this end, we propose \textbf{ClueWeaver}, a dual-agent pipeline for compact local
models in literary and humanities-oriented long-narrative reading. Rather than
compressing a long document into a single latent state, \textbf{ClueWeaver} builds
retrieval-aware narrative segments and uses a Finder to select passages
with answer-critical clues. These passages are packed with paragraph IDs and
narrative order preserved, giving the Interpreter readable evidence for
connecting clues, answering the question, and, when needed, applying
Interpreter$_{\mathrm{self\text{-}cal}}$ as a final consistency check.
This separation makes failures easier to localize across candidate construction,
clue selection, evidence packing, and final reasoning. We further optimize both
agents with reward-guided training: the Finder is rewarded for retaining
supporting clues and faithful paragraph references, while the
Interpreter is rewarded for correct, grounded, and compact rationales.
In this setting, reward-guided reinforcement learning provides a practical way
to align intermediate evidence decisions with task-level outcomes, consistent
with recent progress in feedback- and reward-based post-training for instruction
following and reasoning~\cite{shao2024deepseekmath}.
Our contributions are three-fold:
\begin{enumerate}
\setlength{\topsep}{0pt}
\setlength{\partopsep}{0pt}
\setlength{\parsep}{0pt}
\setlength{\itemsep}{0.1em}
    \item We propose \textbf{ClueWeaver}, an evidence-aware agentic pipeline that
    decomposes long-narrative question answering into explicit evidence
    selection, self-calibrated interpretation, and evidence-grounded explanation
    for compact local models.
    \item We optimize both the Finder and the Interpreter with
    reward-guided reinforcement learning, encouraging high-recall evidence
    retention, faithful paragraph referencing, and reliable answer generation.
    \item Experiments across multiple long-context narrative understanding
    settings show that \textbf{ClueWeaver} substantially improves compact locally
    deployable language models while providing inspectable evidence traces.
\end{enumerate}

\section{Related Work}

\subsection{Agentic-Assisted Literary and Long-Context Reading}

Literary long-context reading requires tracking plot, characters, temporal
order, and implicit causality. NarrativeQA~\cite{kocisky2018narrativeqa} and
QuALITY~\cite{pang2022quality} introduced QA over books, scripts, and long
inputs; NovelQA~\cite{wang2025novelqa}, DetectiveQA~\cite{xu2025detectiveqa},
and NoCha~\cite{karpinska2024one} extend this line to novel-scale QA and claim
verification. General long-context benchmarks such as
LongBench~\cite{bai2024longbench}, $\infty$Bench~\cite{zhang2024inftybench},
and LongBench v2~\cite{bai2025longbenchv2} further
show that larger windows alone do not guarantee reliable use of buried evidence
inside long inputs~\cite{liu2024lost}. These benchmarks expose the difficulty
of long-range understanding, but mostly evaluate final answers rather than
explicit clue discovery and evidence preservation.

\subsection{Agentic RAG and LLM Reasoning Pipelines}

Retrieval-Augmented Generation (RAG)~\cite{lewis2020retrieval} grounds
generation in retrieved passages. Agentic variants add adaptive control:
ReAct~\cite{yao2023react} couples reasoning with actions,
FLARE~\cite{jiang2023active} retrieves during generation, and
Self-RAG~\cite{asai2024selfrag} adds retrieval and self-critique; related
decision-based search agents have also been studied for knowledge-based visual
QA~\cite{chen2026learning}. Multi-step systems, including
IRCoT~\cite{trivedi2023interleaving}, Self-Ask~\cite{press2023measuring},
Chain-of-Agents~\cite{zhang2024chain}, and OPERA~\cite{liu2026opera}, structure
retrieval and reasoning into staged procedures, while
RAG-DDR~\cite{li2025rag} optimizes RAG with data-driven rewards. These methods
mainly address open-domain retrieval, multi-hop QA, or broad long-context
aggregation. ClueWeaver instead targets closed-document narratives, where the
source is fixed and the key problem is retaining sparse story clues with
paragraph-level citations before final reasoning.

\subsection{Reward-Guided Reinforcement Learning for Reasoning}

Post-training improves model behavior through instruction tuning
SFT~\cite{wei2022finetuned}, preference optimization with PPO-based
RLHF~\cite{schulman2017proximal}, and DPO~\cite{rafailov2023direct}.
DeepSeekMath~\cite{shao2024deepseekmath} shows that GRPO can strengthen
reasoning, and CRAFT~\cite{liu2026craft} uses RL to improve answer-faithful
traces. Answer-attribution analysis~\cite{wang2025reasoningretrieval} further
shows that answers may mix explicit reasoning with retrieval-like memorized
knowledge. In contrast, ClueWeaver applies reward-guided training directly to
paragraph-level evidence selection, citation fidelity, and grounded explanation
for long-narrative reading.

\section{Methodology}

\begin{figure}[t]
\centering
\includegraphics[width=1\textwidth]{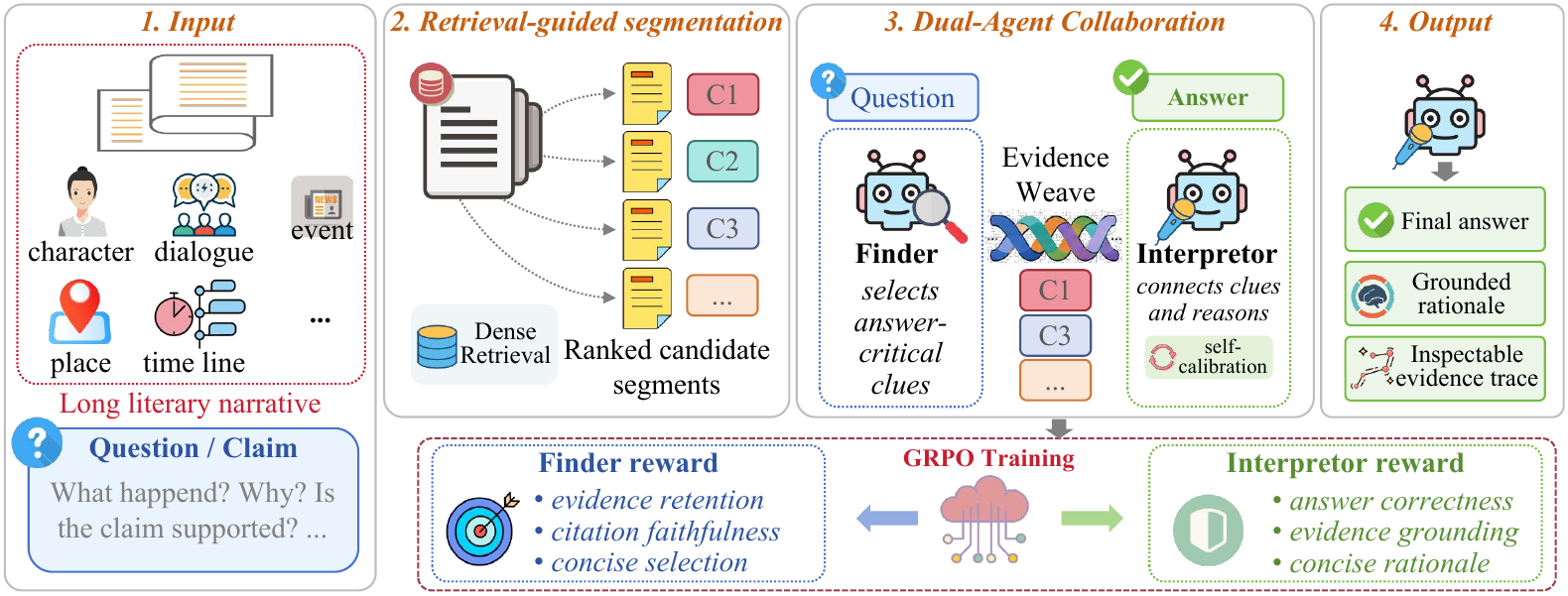}
\caption{Overview of ClueWeaver: Finder selects evidence-bearing narrative
segments, and Interpreter produces a grounded answer with self-calibration.}
\label{fig:overview}
\end{figure}

\subsection{Problem Definition}

We formulate agentic support for long-narrative reading as question answering with
compact language models for resource-constrained humanities settings. Let
$X=\{p_1,p_2,\ldots,p_m\}$ be a narrative, where each $p_i$ is an indexed
paragraph and $m$ is the number of paragraphs. Given a question or
claim $q$, the model predicts answer $y$ using only $X$; $y$ may be a
multiple-choice option or a binary verification label. Since only a small part
of $X$ supports $y$, the task is to select a compact evidence set
$E\subset X$, preserve its narrative order, and generate both $y$ and an
explanation whose claims can be traced to paragraph-referenced evidence in $E$.

\subsection{Overview}

Figure~\ref{fig:overview} illustrates the pipeline, and
Algorithm~\ref{alg:clueweaver} gives the inference procedure. ClueWeaver does not
ask a compact model to read the whole narrative in one pass. It first builds
retrieval-aware segments, uses the \textbf{Finder} to keep clue-bearing passages with
paragraph IDs, packs the retained evidence in narrative order, and lets the
\textbf{Interpreter} produce a paragraph-grounded answer. For binary claims and high-risk question
forms, the same \textbf{Interpreter} optionally runs
$I_{\theta_i,\mathrm{self\text{-}cal}}$, a self-calibration mode that re-checks the
provisional answer against the identical evidence packet.
In Algorithm~\ref{alg:clueweaver}, $\mathcal{Y}$ is the answer space, $B$ is the
evidence budget, $\theta_f$ and $\theta_i$ are the parameters of the two agents,
$\mathcal{S}$ is the segment set, $E$ is the final evidence packet, and
$\tau(q)$ triggers self-calibration. In the implementation, both agents use a
compact XML interface: \texttt{<reason>} contains the paragraph-referenced explanation and
\texttt{<answer>} contains either a YES/NO decision or the final answer.
Complete prompts are in Appendix E.

\begin{algorithm}[t]
\caption{ClueWeaver Inference}
\label{alg:clueweaver}
\begin{algorithmic}[1]
\Require Narrative $X=\{p_i\}_{i=1}^{m}$, question or claim $q$, answer space
$\mathcal{Y}$, evidence budget $B$
\Ensure XML-style output $z^I=(r,y,\mathcal{C})$, with paragraph references extracted from $r$
\State $\mathcal{S}=\{s_j\}_{j=1}^{n}\leftarrow \mathrm{Segment}(X,q)$
\Comment{$n$ is the number of retrieval-aware segments}
\State Initialize candidate evidence pool $C\leftarrow \emptyset$
\For{$j=1$ to $n$}
    \State \textbf{Finder}: $z_j^F\leftarrow F_{\theta_f}(s_j,q)$
    \Statex \hspace{\algorithmicindent}
    $z_j^F=
    \begin{array}[t]{l}
    \langle\texttt{reason}\rangle u_j \langle/\texttt{reason}\rangle\\
    \langle\texttt{answer}\rangle d_j \langle/\texttt{answer}\rangle
    \end{array}$
    \State Parse $z_j^F$ into $(u_j,d_j)$ and extract referenced paragraph IDs $e_j$
    \If{$d_j=\mathrm{YES}$}
        \State $C\leftarrow C\cup \{(s_j,e_j,u_j)\}$
    \EndIf
\EndFor
\State Order $C$ by the paragraph indices inherited from $X$
\State Build $E=\mathrm{Pack}(C,B)$ by keeping ordered clues within budget $B$
\State \textbf{Interpreter}: $\hat{z}^I\leftarrow I_{\theta_i,\mathrm{ans}}(E,q,\mathcal{Y})$
\Statex \hspace{\algorithmicindent}
$\hat{z}^I=
\begin{array}[t]{l}
\langle\texttt{reason}\rangle r \langle/\texttt{reason}\rangle\\
\langle\texttt{answer}\rangle y \langle/\texttt{answer}\rangle
\end{array}$
\If{$\tau(q)=1$}
    \State $z^I\leftarrow I_{\theta_i,\mathrm{self\text{-}cal}}(E,q,\mathcal{Y},\hat{z}^I)$
\Else
    \State $z^I\leftarrow \hat{z}^I$
\EndIf
\State Parse $z^I$ into $(r,y)$ and extract paragraph references $\mathcal{C}$ from $r$
\State \Return $z^I=(r,y,\mathcal{C})$
\end{algorithmic}
\end{algorithm}

\subsection{Training Principle}

We train the two agents with Group Relative Policy Optimization
(GRPO)~\cite{shao2024deepseekmath}. Let $a\in\{F,I\}$ denote the agent, where
$F$ is the \textbf{Finder} and $I$ is the \textbf{Interpreter}. The Finder input is
$x^F=(s_j,q)$ with output $o^F=z_j^F$, while the Interpreter input is
$x^I=(E,q,\mathcal{Y})$ with output $o^I=z^I$. For each input $x^a$, the old
policy $\pi^a_{\mathrm{old}}$ samples $K$ candidate outputs
$\{o_k^a\}_{k=1}^{K}$. Each output is scored by the agent-specific reward
$R_a(o_k^a;x^a)$, and its group-normalized advantage is
\begin{equation}
\label{eq:grpo-advantage}
A_k^a=
\frac{R_a(o_k^a;x^a)-\mathrm{mean}_{l}R_a(o_l^a;x^a)}
{\mathrm{std}_{l}R_a(o_l^a;x^a)+\epsilon},
\end{equation}
where $\epsilon$ is a small numerical constant. Let
$\rho_k^a=\pi_{\theta_a}^a(o_k^a|x^a)/
\pi_{\mathrm{old}}^a(o_k^a|x^a)$ and
$\bar{\rho}_k^a=\mathrm{clip}(\rho_k^a,1-\delta,1+\delta)$. The clipped objective
for agent $a$ is
\begin{equation}
\label{eq:grpo-objective}
\begin{aligned}
\mathcal{J}_{a}(\theta_a)
&=\mathbb{E}_{x^a,\,\mathbf{o}^a\sim\pi^a_{\mathrm{old}}}
\Bigg[
\frac{1}{K}\sum_{k=1}^{K}
\Big(
\min(\rho_k^a A_k^a,\bar{\rho}_k^a A_k^a)
-\beta D_k^a
\Big)
\Bigg],
\end{aligned}
\end{equation}
where $\theta_a$ is the trainable policy parameter, $\delta$ is the clipping
range, $\beta$ controls KL regularization, and the sample-level KL estimator is
\begin{equation}
\label{eq:grpo-kl}
D_k^a
=
\frac{\pi_{\mathrm{ref}}^a(o_k^a|x^a)}{\pi_{\theta_a}^a(o_k^a|x^a)}
-\log
\frac{\pi_{\mathrm{ref}}^a(o_k^a|x^a)}{\pi_{\theta_a}^a(o_k^a|x^a)}
-1 .
\end{equation}
This relative objective is useful for ClueWeaver because many valid rationales
can exist for the same narrative question, while their usefulness can still be
judged by task-level signals. We therefore use the same optimization form for both agents but instantiate agent-specific rewards. $R_F$ favors high-recall
clue retention, faithful paragraph-ID references, and calibrated YES/NO decisions,
giving much higher reward to retaining answer-bearing evidence than to rejecting
extra candidates. $R_I$ favors answer correctness, paragraph-grounded support, concise
explanation, and resistance to unsupported inference. Thus, training directly
aligns the pipeline stages: preserving answer-critical evidence before reasoning
and converting the retained evidence into a grounded final answer.

\subsection{Finder: Evidence Selection and Rationale Generation}

The \textbf{Finder} is responsible for converting a long narrative into a small
set of answer-relevant evidence with short rationales. Given the paragraph
sequence $X$ and the question $q$, we first build retrieval-aware segments
$\mathcal{S}=\{s_j\}_{j=1}^{n}$. Each
segment $s_j$ contains a contiguous paragraph span $I_j\subseteq\{1,\ldots,m\}$.
The segmentation uses lexical and dense retrieval scores to place short anchor
segments around paragraphs that are likely to be relevant to $q$, while the
remaining text is covered by local windows. Retrieval is therefore used to guide
segmentation boundaries, not to replace reading of the given narrative. This
keeps the input to the \textbf{Finder} short enough for a compact model, while retaining
paragraph indices needed for later evidence tracing.
For each segment $s_j$, the \textbf{Finder} predicts
\begin{equation}
\label{eq:finder-output}
(d_j,e_j,u_j)=F_{\theta_f}(s_j,q),
\end{equation}
where $d_j\in\{\mathrm{YES},\mathrm{NO}\}$ is the clue decision,
$e_j\subseteq I_j$ lists referenced paragraph IDs, and $u_j$ is the rationale.
In the XML output, $u_j$ is written in
\texttt{<reason>}, $d_j$ in \texttt{<answer>}, and $e_j$ is extracted from the
paragraph-ID references inside $u_j$. Segments with $d_j=\mathrm{YES}$ are added to
the candidate clue pool $C$. Since long-narrative questions often depend on
indirect or distributed clues, the \textbf{Finder} is designed as a high-recall selector:
it should avoid discarding answer-supporting evidence even when the segment does
not directly state the final answer.

\noindent\textbf{RL Training for Finder.}
We train the \textbf{Finder} with reward-guided reinforcement learning over structured
outputs. For a training segment, let $z_j\in\{0,1\}$ be the gold evidence label
and let $G_j\subseteq I_j$ be the annotated supporting paragraphs within the
segment. The \textbf{Finder} reward combines decision and evidence behavior:
\begin{equation}
\label{eq:finder-reward}
R_F =
\lambda_{\mathrm{fmt}}R_{\mathrm{fmt}}
+\lambda_{\mathrm{dec}}R_{\mathrm{dec}}
+\lambda_{\mathrm{cite}}R_{\mathrm{cite}}
+\lambda_{\mathrm{comp}}R_{\mathrm{comp}}
+\lambda_{\mathrm{neg}}R_{\mathrm{neg}} .
\end{equation}
Here each $\lambda$ is a non-negative weight controlling the importance of its
corresponding reward term. $R_{\mathrm{fmt}}$ rewards valid structured output,
$R_{\mathrm{dec}}$ rewards the correct YES/NO decision, and
$R_{\mathrm{cite}}$ measures overlap between predicted paragraph IDs $e_j$ and gold
paragraphs $G_j$. $R_{\mathrm{comp}}$ rewards compact gold references, and
$R_{\mathrm{neg}}$ rewards concise \textsc{NO} rationales without unsupported
IDs. A missed positive segment receives only the format reward. This weighted reward design reflects the role of
the \textbf{Finder} in the pipeline: preserving answer-critical clues is more
important than over-filtering the narrative, since the
\textbf{Interpreter} can only reason from retained evidence. See Appendix A for
details.

\subsection{Interpreter: Evidence-Grounded Interpretation}

\noindent\textbf{Evidence-Grounded Interpretation.}
The \textbf{Interpreter} turns the selected evidence packet into the final
answer. Let
$E=\{(\ell_t,\tilde{p}_t,\tilde{u}_t)\}_{t=1}^{T}$ denote the ordered packet
after packing, where $\ell_t$ is the original paragraph index, $\tilde{p}_t$ is
the retained evidence text, $\tilde{u}_t$ is the \textbf{Finder} rationale, and
$T$ is the number of packed evidence units. The \textbf{Interpreter} first predicts a
provisional answer and then, when triggered, self-calibrates it using the same
evidence:
\begin{equation}
\label{eq:interpreter-output}
\begin{aligned}
\hat{z}^{I} &= I_{\theta_i,\mathrm{ans}}(E,q,\mathcal{Y}),\\
z^{I} &=
\begin{cases}
I_{\theta_i,\mathrm{self\text{-}cal}}(E,q,\hat{z}^{I}), & \tau(q)=1,\\
\hat{z}^{I}, & \tau(q)=0,
\end{cases}
\end{aligned}
\end{equation}
where $z^I$ is parsed into $(r,y)$ and referenced paragraphs $\mathcal{C}$. Here $r$
is a concise evidence-grounded rationale written in \texttt{<reason>},
$y\in\mathcal{Y}$ is the final answer written in \texttt{<answer>}, and
$\mathcal{C}\subseteq\{\ell_t\}_{t=1}^{T}$ is extracted from paragraph-ID references
inside $r$. The
trigger $\tau(q)$ is active for binary claim verification and for multiple-choice
questions whose wording suggests higher risk of polarity or reasoning errors,
such as negation, exception, causal, or inferential forms. The self-verifier is
internal to the \textbf{Interpreter}; it re-checks the provisional answer against the
same evidence packet, uses the same compact model, and does not introduce a third
agent. The model is therefore not asked to freely summarize the whole narrative.
It must connect the selected clues, choose an answer from the allowed answer
space, and make the rationale traceable to paragraph IDs.

\noindent\textbf{RL Training for Interpreter.}
We train the \textbf{Interpreter} with the same GRPO principle but a different
target. Given gold answer $y^\star$ and, when available, the supplied evidence
paragraph set $H$, the \textbf{Interpreter} uses
\begin{equation}
\label{eq:interpreter-reward}
R_I =
\lambda_{\mathrm{fmt}}R_{\mathrm{fmt}}
+\lambda_{\mathrm{ans}}R_{\mathrm{ans}}
+\lambda_{\mathrm{cite}}R_{\mathrm{cite}}
+\lambda_{\mathrm{ground}}R_{\mathrm{ground}}
-\lambda_{\mathrm{hall}}R_{\mathrm{hall}} .
\end{equation}
Here $R_{\mathrm{fmt}}$ rewards valid structured output,
$R_{\mathrm{ans}}$ rewards matching the gold answer $y^\star$,
$R_{\mathrm{cite}}$ rewards paragraph IDs that point to $H$ and penalizes
invalid IDs, and $R_{\mathrm{ground}}$ rewards concise rationales with concrete
grounding signals. The hallucination penalty $R_{\mathrm{hall}}$ discourages
unsupported uncertainty or invalid paragraph references. This makes the \textbf{Interpreter}
conservative with evidence: it is rewarded for correctness and traceability.

\section{Experiments}

\subsection{Experimental Setup}

\noindent\textbf{Datasets.}
We evaluate on four long-context narrative settings using the same test
instances for all methods: DetectiveQA~\cite{xu2025detectiveqa} for sparse
detective-plot clues, $\infty$Bench~\cite{zhang2024inftybench} and LongBench
v2~\cite{bai2025longbenchv2} for broader long-context reasoning, and
NoCha~\cite{karpinska2024one} for novel-length claim verification.

\noindent\textbf{Baselines.}
We compare with two groups. End-to-end readers use the available context
directly with naive head truncation, including the
$4$B backbone and larger local models: Qwen3-8B~\cite{yang2025qwen3},
Ministral-3-14B~\cite{mistral2025ministral3}, GPT-OSS-20B~\cite{openai2025gptoss},
Qwen3-30B-A3B~\cite{yang2025qwen3}, and
Gemma-4-31B-it~\cite{google2026gemma4}. We also report higher-cost API readers,
Claude Haiku 4.5~\cite{anthropic2025haiku45} and GPT-5 nano~\cite{openai2025gpt5developers},
as large-context references. Agentic baselines include ReAct~\cite{yao2023react},
IRCoT~\cite{trivedi2023interleaving}, Self-Ask~\cite{press2023measuring},
Chain-of-Agents~\cite{zhang2024chain}, and RAG-DDR~\cite{li2025rag}, all using
BGE-M3~\cite{chen2024m3embedding}.

\noindent\textbf{Metrics.}
We report final answer accuracy, normalizing multiple-choice outputs to option
labels and verification outputs to binary labels.

\noindent\textbf{Implementation Details.}
To control local-model comparisons, all local end-to-end readers use the same
$32$K setting: maximum context length $32{,}768$, tokenizer-exact input budget
$30{,}592$, and $128$ output tokens. For higher-cost API LLMs, we report a
separate large-context setting with a $128$K budget, using $126{,}976$ input
tokens and the same $128$-token output cap. ClueWeaver, the agentic baselines,
and the $4$B end-to-end reader use Qwen3-4B-Instruct~\cite{yang2025qwen3};
larger local end-to-end readers use their own weights. Retrieval-based methods
share BGE-M3~\cite{chen2024m3embedding} and the same answer parser; baseline,
training, and implementation details are deferred to Appendix C.

\begin{table}[t]
\centering
\caption{Main results on four long-context narrative benchmarks (final answer
accuracy, \%). Among local methods, best per column is in \textbf{bold} and
second best is \underline{underlined}; best API result per column is marked with
\protect\uwave{wavy underlines}.}
\label{tab:main}
\setlength{\tabcolsep}{4pt}
\begin{tabular}{lccccc}
\toprule
Method & DetectiveQA & $\infty$Bench & LongBench v2 & NoCha & Overall \\
\midrule
\rowcolor{typeGray}
\multicolumn{6}{l}{\textit{End-to-end reader (Local)}} \\
Qwen3-4B-Instruct~\cite{yang2025qwen3}  & 36.5 & 50.7 & 38.5 & 49.5 & 44.5 \\
Qwen3-8B~\cite{yang2025qwen3}           & 45.2 & 56.5 & 26.9 & 55.0 & 49.7 \\
Ministral-3-14B~\cite{mistral2025ministral3} & 45.2 & \underline{60.9} & 26.9 & 51.4 & 49.4 \\
GPT-OSS-20B~\cite{openai2025gptoss}      & 30.8 & 33.3 & 34.6 & 50.5 & 38.7 \\
Qwen3-30B-A3B~\cite{yang2025qwen3}      & 44.2 & 58.0 & 38.5 & 54.1 & 50.3 \\
Gemma-4-31B-it~\cite{google2026gemma4}  & 35.6 & 58.0 & \underline{46.2} & 59.5 & 50.0 \\
\midrule
\rowcolor{typeGray}
\multicolumn{6}{l}{\textit{End-to-end reader (API)}} \\
Claude Haiku 4.5~\cite{anthropic2025haiku45} & \uwave{62.5} & 73.9 & \uwave{38.5} & \uwave{64.9} & \uwave{63.9} \\
GPT-5 nano~\cite{openai2025gpt5developers} & \uwave{62.5} & \uwave{76.8} & 26.9 & 60.4 & 61.9 \\
\midrule
\rowcolor{typeGray}
\multicolumn{6}{l}{\textit{Agentic pipelines}} \\
ReAct~\cite{yao2023react}              & 50.0 & 50.7 & 38.5 & 58.6 & 52.3 \\
IRCoT~\cite{trivedi2023interleaving}           & \underline{53.8} & 44.9 & 34.6 & \underline{60.4} & \underline{52.6} \\
Self-Ask~\cite{press2023measuring}        & 36.5 & 49.3 & 19.2 & 59.5 & 46.1 \\
Chain-of-Agents~\cite{zhang2024chain} & 27.9 & 46.4 & 38.5 & 55.9 & 42.9 \\
RAG-DDR~\cite{li2025rag}         & 47.1 & 53.6 & 30.8 & 59.5 & 51.6 \\
\midrule
\rowcolor{typeGray}
\multicolumn{6}{l}{\textit{Our method}} \\
\textbf{ClueWeaver (ours)} & \textbf{55.8} & \textbf{63.8} & \textbf{50.0} & \textbf{61.3} & \textbf{59.0} \\
\textit{$\Delta$ vs.\ best local} & $+2.0$ & $+2.9$ & $+3.8$ & $+0.9$ & $+6.4$ \\
\textit{$\Delta$ vs.\ best API} & $-6.7$ & $-13.0$ & $+11.5$ & $-3.6$ & $-4.9$ \\
\bottomrule
\end{tabular}
\end{table}

\subsection{Main Results}

Table~\ref{tab:main} reports final answer accuracy on the four benchmarks.
ClueWeaver achieves the best local overall accuracy ($59.0\%$), leads all local
methods on every dataset, and improves over the strongest local baseline (IRCoT,
$52.6\%$) by $+6.4$ points overall. With the same Qwen3-4B backbone, the direct end-to-end reader reaches only $44.5\%$ because
many narratives still require truncation; ClueWeaver improves it by $+14.5$
points. The gain is not a scale effect: the best end-to-end reader up to $31$B
(Qwen3-30B-A3B, $50.3\%$) remains $8.7$ points behind ClueWeaver's $4$B
backbone. These results show that retrieval-aware evidence selection, rather
than context length or model size alone, is central to compact local
long-narrative QA. Higher-cost API LLMs are reported as large-context references;
ClueWeaver trails the best API overall result by $4.9$
points and surpasses it on LongBench v2 by $11.5$ points.

\subsection{Ablation Study}

We ablate ClueWeaver on DetectiveQA, whose sparse, distributed clues make the contribution of each part most visible.

\noindent\textbf{Component ablation.}
Table~\ref{tab:ablation}(a) removes inference components from the full model.
Disabling \textbf{Interpreter}$_{\mathrm{self\text{-}cal}}$ lowers accuracy by
$4.8$ to $51.0\%$, showing that the internal second pass helps correct fragile
decisions over the same evidence. Removing the Finder---dumping all retrieved
passages to the Interpreter instead of selecting answer-critical evidence---drops
accuracy by  $5.8$ points to $50.0\%$. Further removing both agents, leaving a bare
end-to-end reader that must truncate the narrative to the backbone window, falls
to $36.5\%$.
Thus, selected evidence and self-calibration both contribute to reliable
long-narrative reading.

\begin{table}[t]
\centering
\caption{Ablation on DetectiveQA. (a) pipeline components and (b) RL training are
removed from the full model ($\Delta$: accuracy change vs.\ the full model).}
\label{tab:ablation}
\begin{minipage}[t]{0.49\textwidth}
\centering
(a) Component ablation\\[2pt]
\begin{tabular}{lcc}
\toprule
Configuration & Acc. & $\Delta$ \\
\midrule
ClueWeaver (full) & \textbf{55.8} & -- \\
\quad w/o Interpreter$_{\mathrm{self\text{-}cal}}$ & 51.0 & $-4.8$ \\
\quad w/o Finder & 50.0 & $-5.8$ \\
\quad w/o Finder \& Interpreter & 36.5 & $-19.3$ \\
\bottomrule
\end{tabular}
\end{minipage}
\hfill
\begin{minipage}[t]{0.49\textwidth}
\centering
(b) Training ablation\\[2pt]
\begin{tabular}{lcc}
\toprule
Configuration & Acc. & $\Delta$ \\
\midrule
ClueWeaver (full) & \textbf{55.8} & -- \\
\quad w/o Finder RL & 49.0 & $-6.8$ \\
\quad w/o Interpreter RL & 54.8 & $-1.0$ \\
\quad w/o both RL & 50.0 & $-5.8$ \\
\bottomrule
\end{tabular}
\end{minipage}
\end{table}
\begin{figure}[t]
\centering
\includegraphics[width=0.92\textwidth]{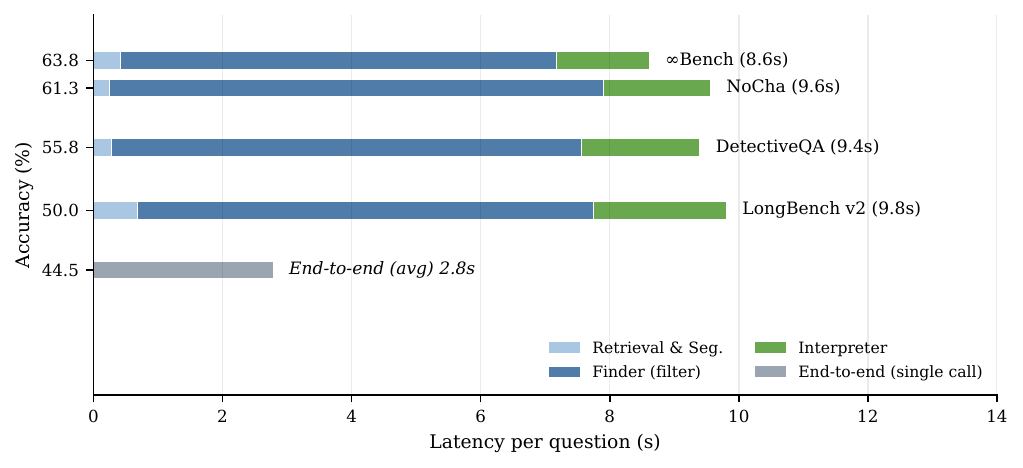}
\caption{Accuracy--latency trade-off using Table~\ref{tab:main} accuracies and
serial single-GPU stage timings. Most additional latency comes from the Finder
stage.}
\label{fig:eff}
\end{figure}

\noindent\textbf{Training ablation.}
Table~\ref{tab:ablation}(b) removes reward-guided RL from the full model with the
pipeline fixed. Removing Finder RL (untrained Finder, RL Interpreter) is most
damaging,  dropping accuracy by $6.8$ points to $49.0\%$---\emph{below} the $50.0\%$ obtained with no
Finder (Table~\ref{tab:ablation}a): an untrained Finder discards useful
evidence, so it is RL training that turns the Finder into a net gain. Removing
Interpreter RL costs $1.0$; removing both returns to the untrained pipeline
at $50.0\%$. Finder training contributes more by retaining
answer-critical clues, while Interpreter training converts the selected evidence
into correct answers.

\begin{table}[t]
\centering
\caption{Cost profile for local and API readers.}
\label{tab:cost}
\scriptsize
\setlength{\tabcolsep}{2pt}
\begin{tabular}{@{}p{0.28\linewidth}cp{0.18\linewidth}p{0.30\linewidth}@{}}
\toprule
Setting & Overall Acc. & Access & Example GPU \\
\midrule
\rowcolor{typeGray}
\multicolumn{4}{l}{\textit{End-to-end reader (Local)}} \\
Qwen3-4B~\cite{yang2025qwen3} & 44.5 & 16GB & RTX 5060 Ti 16GB \\
Qwen3-8B~\cite{yang2025qwen3} & 49.7 & 24GB & RTX 4090 / RTX 5090 \\
Ministral-3-14B~\cite{mistral2025ministral3} & 49.4 & 24GB & RTX 4090 / RTX 5090 \\
GPT-OSS-20B~\cite{openai2025gptoss} & 38.7 & 24GB & RTX 4090 / RTX 5090 \\
Qwen3-30B-A3B~\cite{yang2025qwen3} & 50.3 & 80GB & A100/H100  \\
Gemma-4-31B-it~\cite{google2026gemma4} & 50.0 & 80GB & A100/H100  \\
\midrule
\rowcolor{typeGray}
\multicolumn{4}{l}{\textit{Our method}} \\
ClueWeaver & 59.0 & 24GB & RTX 4090 / RTX 5090 \\
\midrule
\rowcolor{typeGray}
\multicolumn{4}{l}{\textit{End-to-end reader (API)}} \\
GPT-5 nano & 61.9 & API & $\approx$\$2/310 \\
Claude Haiku 4.5 & 63.9 & API & $\approx$\$40/310 \\
\bottomrule
\end{tabular}
\end{table}
\subsection{Analysis}

\noindent\textbf{Efficiency.}
When served on a single GPU with Qwen3-4B, ClueWeaver answers in $8.6$--$9.8$\,s per question,
compared with $2.8$\,s for direct reading; most extra latency comes from Finder
calls. The ClueWeaver end-to-end latency remains practical for local deployment, while
bringing a $+14.5$-point gain and evidence-level inspection.

\begin{figure}[t]
\centering
\includegraphics[width=0.95\textwidth]{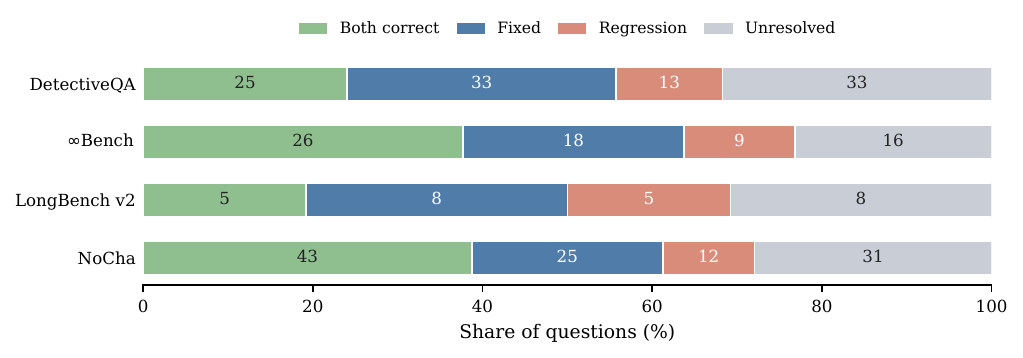}
\caption{Error transitions between the direct reader and ClueWeaver across
the four benchmarks. ClueWeaver yields a clear net correction gain.}
\label{fig:errmat}
\end{figure}
\begin{table}[ht!]
\centering
\caption{Case study on DetectiveQA. ClueWeaver exposes the evidence-selection and
interpretation process, with self-calibration used as a final consistency check.}
\label{tab:case}
\sffamily\small
\renewcommand{\arraystretch}{1.2}
\begin{tabular}{@{}p{0.95\textwidth}@{}}
\toprule
\rowcolor{caseQ}
\textbf{\textsf{Question.}} The reason why the robot doll was pushed downstairs is:\quad

(A)~Break it for study\quad(B)~Eliminate clues\quad(C)~Hurt him\quad(D)~No particular purpose
\hfill\textbf{[Gold: B]}\\
\midrule
\rowcolor{caseBad}
\textbf{\textsf{End-to-end reader}} 
selects the surface-level distractor:\quad
{\ttfamily D}~~\textcolor{badRed}{$\times$}\\
\midrule
\rowcolor{caseGood}
\textbf{\textsf{ClueWeaver}} keeps the decisive evidence and then self-calibrates
the Interpreter output.\\
\rowcolor{caseGood}
\;\;\textbf{\textsf{Finder}} (retained evidence from the log):\newline
{\ttfamily\scriptsize YES segments include [1203], where the killer is described
as wiping physical traces, and [1326], where Dr. Phil links this to pushing the
robot doll downstairs.}\\
\rowcolor{caseGood}
\;\;\textbf{\textsf{Interpreter}}:\newline
{\ttfamily\scriptsize <reason>Option (B) is supported by [1203], where the killer
erases footprints, fingerprints, and dust traces; [1326] links this reasoning to
why the robot doll was pushed downstairs.</reason>

<answer>B</answer>}\\
\rowcolor{caseGood}
\;\;\textbf{\textsf{Interpreter}$_{\mathrm{self\text{-}cal}}$}:\newline
{\ttfamily\scriptsize <reason>The two cited passages are consistent: [1203]
establishes the motive of removing physical traces, and [1326] applies that
motive to the pushed doll. Thus the answer is eliminating clues.
</reason>

<answer>B</answer>}~~\textcolor{okGreen}{$\checkmark$}\\
\bottomrule
\end{tabular}
\end{table}

\noindent\textbf{Cost.}
Table~\ref{tab:cost} compares local GPU requirements with API token fees. Strong
API readers can be competitive, but long-context calls are costly. ClueWeaver
reaches API-level accuracy with a compact local model and can be deployed on
commercial-grade GPUs, making the cost practical for sustained use. Its
deployment footprint is also far below that of 30B-scale local readers.

\noindent\textbf{Error analysis.}
Figure~\ref{fig:errmat} compares direct-reader and ClueWeaver correctness.
ClueWeaver recovers $84$ of $172$ direct-reader errors, yielding a consistently
positive net gain across the benchmarks. Some difficult cases remain, especially
when the answer depends on distant multi-hop clues or passages with weak surface
overlap, which points to stronger distant-clue retrieval as a future direction.

\subsection{Case Study}
Table~\ref{tab:case} contrasts direct reading with the full ClueWeaver pipeline.
The end-to-end reader selects a surface-level distractor, whereas ClueWeaver first
finds the relevant paragraph, then grounds the answer in that evidence. The
self-calibration step appears as a final consistency check inside the
\textbf{Interpreter}, not as a separate agent. Additional qualitative cases are
shown in Appendix D.

\section{Conclusion}
This paper presented ClueWeaver, a reward-guided dual-agent framework for long
narrative QA with compact local models. The framework separates evidence
selection from interpretation: Finder retains sparse clue passages, while
Interpreter connects them into grounded answers and applies self-calibration for
high-risk questions. Reward-guided training aligns both stages with the end
task, encouraging clue preservation, faithful paragraph references, and concise
explanations. Experiments show that ClueWeaver improves local end-to-end
readers, outperforms much larger local models, and approaches stronger API
readers while remaining deployable on commercial GPUs. These results suggest
that compact models can support literary and humanities analysis through
explicit evidence paths. Future work will focus on harder distant clues and
more robust multi-hop evidence integration.

\subsubsection*{Acknowledgements}
This research was supported by the National Key R\&D Program of China
(No. 2023YFC3303800). We also acknowledge WisPaper and QiewenPaper for providing
Academic Agent support and GPU computational resources throughout the study; see
\url{https://wispaper.ai}.

\bibliographystyle{splncs04}
\bibliography{ref}

\clearpage
\appendix

\section{Reward Design Details}
\label{app:reward}
\subsection{Reward Components}
We use rule-based rewards rather than a learned reward model, because the two
agents have explicit structured roles and the training signals can be defined
from labels, paragraph indices, and output format. The final trained agents use
the balanced Finder reward and the test-aligned Interpreter reward described
below. All rewards are computed after parsing the generated XML-style output. If
the output cannot be parsed, the reward is set to zero. Otherwise, format
validity provides a small base reward, while task-specific correctness and
evidence behavior determine the remaining score. The constants below are the
unnormalized rewards used before GRPO advantage normalization.

\paragraph{Finder reward.}
For a segment $s_j$, let $z_j\in\{0,1\}$ indicate whether it contains supporting
evidence, let $G_j$ be the annotated gold paragraphs inside the segment, and let
$P_j$ be the paragraph IDs referenced in the Finder's \texttt{<reason>} field. The final Finder
reward is balanced: it makes positive evidence retention more valuable than a
short negative response, while still giving enough reward to correct
\textsc{NO} decisions to prevent an all-\textsc{YES} policy.

\begin{table}[h]
\centering
\caption{Finder reward components.}
\label{tab:finder-reward-details}
\small
\renewcommand{\arraystretch}{1.12}
\begin{tabular}{@{}p{0.24\textwidth}p{0.50\textwidth}p{0.16\textwidth}@{}}
\toprule
Component & Purpose & Score \\
\midrule
Format validity & Output must follow the required structured format and expose
a decision and rationale. & $+1.0$ \\
Positive decision & A gold-bearing segment should be kept by predicting
\textsc{YES}. & $+1.5$ \\
Paragraph-ID F1 & Reward overlap between predicted and gold paragraph IDs,
using F1 over $P_j$ and $G_j$. & up to $+1.5$ \\
Compact ID bonus & Extra credit when the Finder references gold paragraphs
without adding many irrelevant paragraph IDs. & $+0.25$ \\
Correct rejection & A non-evidence segment should be rejected by predicting
\textsc{NO}. & $+1.25$ \\
Negative rationale quality & Reward non-trivial \textsc{NO} rationales with no
paragraph ID and explicit irrelevance cues. & up to $+0.75$ \\
Length calibration & Encourage concise but informative rationales rather than
empty templates or long summaries. & up to $+0.25$ \\
Spurious ID control & Unsupported paragraph IDs on negative segments receive
little or no additional credit. & capped \\
\bottomrule
\end{tabular}
\end{table}

This design intentionally favors recall without collapsing into a trivial
all-\textsc{YES} policy. Positive segments can receive the highest score only
when the decision and paragraph IDs are both correct; negative segments still
receive meaningful reward when they are rejected with a concise explanation.

\paragraph{Interpreter reward.}
The Interpreter receives the packed evidence and predicts the final answer. Its
reward is correctness-dominant, with all rationale-quality bonuses gated by a
correct answer. This prevents the model from receiving high reward for fluent
but wrong explanations.

\begin{table}[t]
\centering
\caption{Interpreter reward components.}
\label{tab:interpreter-reward-details}
\small
\renewcommand{\arraystretch}{1.12}
\begin{tabular}{@{}p{0.24\textwidth}p{0.50\textwidth}p{0.16\textwidth}@{}}
\toprule
Component & Purpose & Score \\
\midrule
Format validity & Output must contain a parseable rationale and normalized
answer field. & $+1.0$ \\
Answer correctness & Reward the normalized final answer. Multiple-choice
questions receive a larger gain than binary verification because their random
baseline is lower. & $+2.5$ MCQ; $+1.5$ binary \\
Hard-case bonus & Additional reward for correctly solving examples marked as
hard residual errors during training. & $+0.5$ \\
Paragraph-ID use & Reward rationales that reference paragraph IDs from the supplied
evidence. Invalid paragraph IDs are penalized when the allowed paragraph set is
known. & $+0.3/-0.3$ \\
Rationale length & Encourage concise but non-trivial rationales rather than
bare answers or long summaries. & $+0.2$ \\
Specificity & Reward concrete grounding signals such as quoted phrases,
numbers, or named entities from the evidence. & $+0.3$ \\
Hedging penalty & Penalize unsupported uncertainty templates such as
``cannot determine'' or ``insufficient evidence''. & $-0.4$ \\
\bottomrule
\end{tabular}
\end{table}

The two rewards therefore optimize complementary abilities. The Finder is pushed
to preserve sparse, answer-relevant clues with faithful paragraph-ID references,
whereas the Interpreter is pushed to convert the retained evidence into a
correct, grounded, and concise answer.

\subsection{GRPO Objective Details}
For each agent $a\in\{F,I\}$ and training input $x^a$, GRPO samples $K$ complete
structured outputs from the previous policy and compares them within the same group.
Let $o_k^a=(w_{k,1},\ldots,w_{k,T_k})$ be the $k$-th sampled token sequence. Its
sequence log-probability under a policy $\pi$ is
\begin{equation}
\label{eq:app-seq-logprob}
\log \pi(o_k^a|x^a)=
\sum_{t=1}^{T_k}\log \pi(w_{k,t}|x^a,w_{k,<t}).
\end{equation}
The sequence-level policy ratio used in Eq.~\ref{eq:grpo-objective} is therefore
\begin{equation}
\label{eq:app-ratio}
\rho_k^a=
\exp\!\left(
\log \pi_{\theta_a}^a(o_k^a|x^a)
-\log \pi_{\mathrm{old}}^a(o_k^a|x^a)
\right).
\end{equation}
After parsing the full XML-style response, we compute the agent-specific reward
$R_a(o_k^a;x^a)$ and normalize it within the sampled group:
\begin{equation}
\label{eq:app-adv}
\mu_a=\frac{1}{K}\sum_{l=1}^{K}R_a(o_l^a;x^a),\quad
\sigma_a=\sqrt{\frac{1}{K}\sum_{l=1}^{K}(R_a(o_l^a;x^a)-\mu_a)^2},
\end{equation}
\begin{equation}
\label{eq:app-adv2}
A_k^a=\frac{R_a(o_k^a;x^a)-\mu_a}{\sigma_a+\epsilon}.
\end{equation}
The KL term is evaluated on the same sampled sequence against the reference
model. With
$r_k^a=\pi_{\mathrm{ref}}^a(o_k^a|x^a)/\pi_{\theta_a}^a(o_k^a|x^a)$, we use
\begin{equation}
\label{eq:app-kl}
D_k^a=r_k^a-\log r_k^a-1 .
\end{equation}
This value-free formulation is used for both agents, so no separate critic or
value model is trained.

\section{Training Details}
\label{app:training}
\subsection{Training Data}
\begin{figure}[t]
\centering
\includegraphics[width=0.95\textwidth]{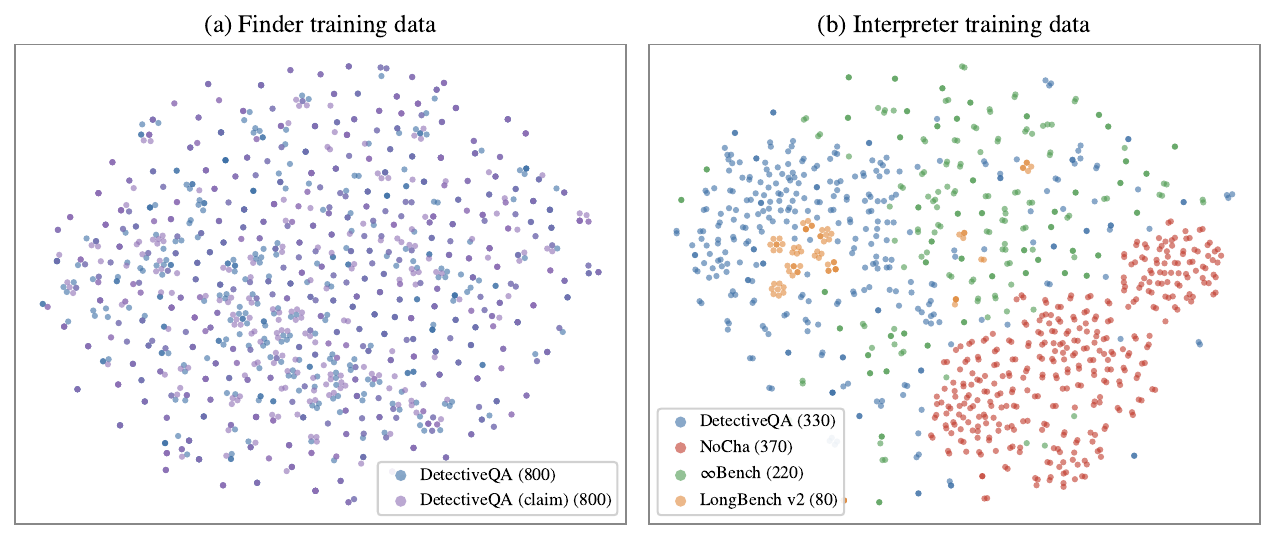}
\caption{t-SNE of BGE-M3 embeddings of the training questions for the two
agents, colored by source. Both pools are detective-domain dominated; the NoCha
portion is mostly synthetic claim-verification, and LongBench v2 is the sparsest
benchmark for the Interpreter.}
\label{fig:datatsne}
\end{figure}

All training data are drawn from the benchmarks' training splits, disjoint from
the test set. As the source narratives are long, often exceeding $100$K tokens,
we segment each document at the paragraph level and supervise both agents on
segments rather than whole texts (Fig.~\ref{fig:datatsne}). The Finder learns
per-segment keep/drop decisions, balanced $50/50$ with hard negatives, from a
question-answering and a claim-verification split of detective novels. The
Interpreter learns from $1{,}000$ examples, each pairing a question with an
evidence packet assembled from selected segments; the mixture follows the test
distribution, scarce real NoCha cases are augmented with synthetic claim
verification, and $27.5\%$ are items the base model fails, concentrating the
reward on hard cases. A t-SNE projection of BGE-M3 question embeddings shows both
pools are dominated by the detective domain; LongBench v2 is the most
under-represented benchmark for the Interpreter, consistent with the smaller
end-to-end gains seen there.

\subsection{Training Setup}
All experiments use Qwen3-4B-Instruct as the base model for both agents. The
Finder and Interpreter are trained separately with GRPO, using eight sampled
responses per prompt to estimate group-relative advantages. We train full model
weights in bfloat16 with cosine learning-rate decay, a warmup ratio of $0.1$,
temperature $1.0$, top-$p=0.9$, top-$k=20$, and GRPO KL coefficient
$\beta=0.04$. We disable model-internal thinking during both training and
inference so that the emitted traces follow the required structured format.

\begin{table}[t]
\centering
\caption{GRPO training hyperparameters for each agent.}
\label{tab:training-config}
\footnotesize
\setlength{\tabcolsep}{4pt}
\renewcommand{\arraystretch}{1.12}
\begin{tabular}{@{}lcc@{}}
\toprule
\textbf{Agent Parameter} & \textbf{Finder} & \textbf{Interpreter} \\
\midrule
base model & Qwen3-4B & Qwen3-4B \\
algorithm & GRPO & GRPO \\
reward & evidence selection & answer grounding \\
lr & $5\times10^{-7}$ & $3\times10^{-7}$ \\
bs & $8$ & $8$ \\
grad\_accum & $8$ & $2$ \\
local\_eff\_bs & $64$ & $16$ \\
global\_eff\_bs & $512$ & $128$ \\
train\_samples & $12$K & $1.0$K \\
max\_steps & $150$ & $100$ \\
group\_sz ($G$) & $8$ & $8$ \\
max\_input\_len & $9216$ & $9216$ \\
max\_output\_len & $256$ & $256$ \\
GRPO beta ($\beta$) & $0.04$ & $0.04$ \\
temp / top-$p$ / top-$k$ & $1.0/0.9/20$ & $1.0/0.9/20$ \\
\bottomrule
\end{tabular}
\end{table}

Training is performed on a single node with 8 x NVIDIA A100 GPUs. During GRPO,
vLLM is colocated with training workers to accelerate rollout generation. At
inference time, each active Qwen3-4B model instance uses about 10--12 GB of GPU
memory with bfloat16 weights. The Finder and Interpreter can therefore be served
sequentially on one GPU or concurrently on separate GPUs. Dense retrieval uses
BGE-M3 on GPU, and all methods share the same answer normalization and output
parser to avoid evaluation differences from formatting alone. Local end-to-end
readers are evaluated with a 32K context budget (30,592 input tokens plus 128
output tokens), while API end-to-end readers use a 128K context budget (126,976
input tokens plus the same 128-token output cap).

\section{Implementation Details}
\label{app:implementation}

\subsection{Baseline Implementations}
\label{app:baselines}
All baselines use the same normalized question format and answer parser as
ClueWeaver. Local agentic baselines use Qwen3-4B-Instruct as the backbone; larger
end-to-end readers use their own model weights. The direct reader receives the
narrative under the configured context budget and answers in one call. BM25,
dense, and hybrid RAG retrieve top paragraphs and answer from the packed
evidence. Our ReAct baseline~\cite{yao2023react} is a ReAct-style iterative RAG
adaptation: in this closed-document setting, the action is retrieval over the
given narrative, the observation is the retrieved paragraph evidence, and the
model alternates reasoning and retrieval before emitting the final answer.
IRCoT~\cite{trivedi2023interleaving} interleaves retrieval with one generated
reasoning sentence per step. Self-Ask~\cite{press2023measuring} first generates
follow-up sub-questions, retrieves evidence for them, and answers from the
resulting trace. Chain-of-Agents~\cite{zhang2024chain} splits the narrative into
chunks, lets worker agents update a communication summary, and uses a manager
agent for the final answer. RAG-DDR~\cite{li2025rag} retrieves candidate
passages, applies a prompt-only knowledge-refinement YES/NO filter, and answers
from the retained passages without DDR training.

\subsection{Retrieval, Segmentation, and Evidence Packing}
\label{app:retrieval-packing}
ClueWeaver uses retrieval as a front-end for candidate construction, not as a
replacement for narrative reading. Given a question $q$ and a long narrative
$D=\{p_i\}_{i=1}^{m}$, we first score paragraphs with dense retrieval and lexical
matching. Dense retrieval is implemented with BGE-M3, while lexical matching is
used only as a complementary signal for robust candidate coverage. The top
paragraphs are used as anchors for local windows, producing candidate segments
that preserve paragraph IDs and nearby context.

Each candidate segment is then judged by the Finder. Unlike standard RAG, which
passes top-ranked chunks directly to the answer model, ClueWeaver asks the
Finder to decide whether a segment contains useful clues and to cite the
supporting paragraph IDs. This step removes many retrieval-only false positives and
keeps evidence traceable at paragraph level.

\begin{table}[t]
\centering
\caption{Evidence construction stages in ClueWeaver.}
\label{tab:evidence-construction}
\small
\renewcommand{\arraystretch}{1.12}
\begin{tabular}{@{}p{0.22\textwidth}p{0.39\textwidth}p{0.29\textwidth}@{}}
\toprule
Stage & Operation & Output \\
\midrule
Retrieval anchors & Score paragraphs using dense retrieval and lexical matching.
Select high-scoring paragraphs as anchors. & Candidate anchor paragraphs \\
Segment construction & Expand each anchor with nearby paragraphs and keep
paragraph IDs. Add local windows to avoid missing surrounding context. &
Ordered candidate segments $s_j$ \\
Finder selection & Predict \textsc{YES}/\textsc{NO}, cite paragraph IDs, and
emit a brief rationale for each segment. & Clue-bearing segments and paragraph IDs \\
Evidence packing & Merge selected evidence, remove duplicates, retain nearby
supporting paragraphs, and sort by original narrative order. & Compact evidence
packet $E$ \\
Interpreter input & Combine $q$, answer options when available, and packed
evidence with paragraph IDs. & Grounded final-answer prompt \\
\bottomrule
\end{tabular}
\end{table}

The evidence packet is controlled by four implementation parameters. $N_E$ is
the maximum number of selected evidence segments. $P_r$ and $P_w$ are the
paragraph budgets for retrieval-anchored and local-window segments, respectively.
$B_c$ is the total character budget for the packed evidence. For example, a
focused configuration uses $N_E=5$, $P_r=2$, $P_w=5$, and $B_c=13{,}000$,
whereas a broader configuration uses $N_E=10$, $P_r=4$, $P_w=6$, and
$B_c=15{,}000$. The former illustrates noise control, while the latter
illustrates recall-oriented packing for sparse and distributed clues. Selected
evidence is sorted by its original paragraph index before being passed to the
Interpreter, so the final model receives clues in narrative order rather than
retrieval-score order.
For the main results, the task-level budgets are DetectiveQA $(10,4,6,15000)$,
$\infty$Bench $(7,3,6,14000)$, LongBench v2 $(8,4,6,15000)$, and NoCha
$(10,6,8,16000)$.

\subsection{Traceability and Format Audit}
\label{app:trace-audit}
We further audit the final ClueWeaver traces used in Table~\ref{tab:main}.
A citation is valid if the paragraph ID cited by the Interpreter appears in the
Finder-provided evidence packet. Missing citations are not counted as invalid;
the audit measures the faithfulness of explicit paragraph references. Across
310 questions, 275 outputs contain paragraph citations. Among citation-bearing
outputs, 685 of 690 row-unique cited paragraph IDs are valid (99.3\%), and 270
of 275 outputs contain only valid citations (98.2\%). The structured output is
also stable: \texttt{<reason>} and \texttt{<answer>} tags are present in 309 of
310 outputs (99.7\%), and the final parser extracts a legal answer label in 308
of 310 outputs (99.4\%).

\begin{table}[t]
\centering
\caption{Citation audit for final ClueWeaver traces. Citation validity is
computed only for outputs with explicit paragraph citations.}
\label{tab:trace-audit}
\small
\setlength{\tabcolsep}{4pt}
\begin{tabular}{lccc}
\toprule
Dataset & Citation-bearing outputs & Valid cited IDs & All citations valid \\
\midrule
DetectiveQA & 98/104 & 235/237 (99.2\%) & 96/98 (98.0\%) \\
$\infty$Bench & 65/69 & 155/155 (100.0\%) & 65/65 (100.0\%) \\
LongBench v2 & 24/26 & 76/76 (100.0\%) & 24/24 (100.0\%) \\
NoCha & 88/111 & 219/222 (98.6\%) & 85/88 (96.6\%) \\
\midrule
Overall & 275/310 & 685/690 (99.3\%) & 270/275 (98.2\%) \\
\bottomrule
\end{tabular}
\end{table}

\section{Additional Case Studies}
\label{app:cases}
Table~\ref{tab:additional-cases} gives representative qualitative examples from
DetectiveQA. We include one fixed case, one regression, and one unresolved case
to show where the proposed pipeline helps and where it still fails.

\begin{table}[t]
\centering
\caption{Additional qualitative cases.}
\label{tab:additional-cases}
\sffamily\footnotesize
\renewcommand{\arraystretch}{1.12}
\begin{tabular}{@{}p{0.18\textwidth}p{0.34\textwidth}p{0.14\textwidth}p{0.24\textwidth}@{}}
\toprule
\rowcolor{caseHeader}
Type & Question & Prediction & Observation \\
\midrule
\rowcolor{caseGood}
\textbf{\textcolor{okGreen}{Fixed error}} &
Henry's wife Sylvia on his drug use situation. &
\textbf{Direct:} \textcolor{badRed}{C}; \textbf{ClueWeaver:} \textcolor{okGreen}{A};
\textbf{Gold:} A &
The Finder retrieves distant paragraphs indicating that Sylvia did not know
about Henry's drug use, allowing the Interpreter to reject plausible but
unsupported alternatives. \\
\midrule
\rowcolor{caseBad}
\textbf{\textcolor{badRed}{Regression}} &
What was the cause of John Fairley's death? &
\textbf{Direct:} \textcolor{okGreen}{D}; \textbf{ClueWeaver:} \textcolor{badRed}{A};
\textbf{Gold:} D &
The selected evidence emphasizes a local suicide explanation and misses the
option-specific detail needed to support the gold answer. \\
\midrule
\rowcolor{caseWarn}
\textbf{Unresolved} &
Tina says: ``The cup is empty.'' What does this mean? &
\textbf{Direct:} \textcolor{badRed}{C}; \textbf{ClueWeaver:} \textcolor{badRed}{C};
\textbf{Gold:} A &
The literal evidence about the empty cup is found, but both systems fail to map
the symbolic statement to the intended narrative implication. \\
\bottomrule
\end{tabular}
\end{table}

\section{Prompt Templates and Structured Outputs}
\label{app:prompts}
We use three prompt families: \textbf{Finder}, \textbf{Interpreter}, and
\textbf{Interpreter}$_{\mathrm{self\text{-}cal}}$. In implementation, each family
has a multiple-choice instantiation and a binary claim-verification
instantiation with \textsc{TRUE}/\textsc{FALSE} answers. Self-calibration is
invoked only inside the \textbf{Interpreter}; there is no self-calibration prompt
for the Finder.

\subsection{Prompt Templates}

\begin{promptbox}{Finder Agent Prompt Template}
You are the evidence Finder agent in a long-narrative QA pipeline.
You read ONE segment from a long story. Paragraphs are numbered like [N].
Decide whether the segment contains concrete evidence that should be shown to a downstream Interpreter.

Say YES only when the segment contains a concrete fact that helps answer the question, choose or rule out one option, or confirm or refute a claim. Strong evidence includes a relevant action, dialogue line, motive, relationship, causal explanation, time/place clue, object, identity, or explicit contradiction.

Say NO when:
- The segment is scene-setting, scenery, weather, or transition narrative.
- The segment mentions characters but says nothing about what the question is asking.
- The overlap is only a common word, option word, or passing name with no relevant fact.
- The segment merely raises suspicion but gives no fact that distinguishes options.
- You cannot name a concrete clue from the segment.

Bias: preserve answer-critical evidence. Prefer NO for pure background, but choose YES for any concrete fact that could help answer, rule out an option, confirm, or refute the claim. Judge this segment independently; partial evidence is still evidence.

Question or claim: {question}

Answer space:
  (A) {opt_a}
  (B) {opt_b}
  (C) {opt_c}
  (D) {opt_d}
  or TRUE/FALSE for claim verification

Segment (paragraphs {start_para}-{end_para}):
{segment_text}

OUTPUT FORMAT -- exactly two XML fields and nothing else:
<reason>one sentence naming the concrete clue, or saying no concrete clue is present; cite [N] when applicable</reason>
<answer>YES</answer> or <answer>NO</answer>
\end{promptbox}

\begin{promptbox}{Interpreter Agent Prompt Template}
You are the final Interpreter in a dual-agent long-narrative QA pipeline.
Answer the question or verify the claim using ONLY the supplied evidence. Evidence paragraphs are numbered like [N].

Rules:
- First identify the exact fact the question asks for; do not drift to a related event.
- Match by meaning, not wording.
- Check question polarity, including false, except, or not true.
- Compare every option against direct evidence.
- Missing evidence for an option does not prove it wrong; only explicit contradiction rules it out.
- Prefer the option with the strongest positive support in the evidence.
- For claim verification, check each essential element of the claim and answer TRUE or FALSE.
- Cite paragraph numbers like [478] for every decisive fact.
- Keep <reason> under 120 words.

Question or claim: {question}

Answer space:
  (A) {opt_a}
  (B) {opt_b}
  (C) {opt_c}
  (D) {opt_d}
  or TRUE/FALSE for claim verification

Evidence (kept segments, in order):
{evidence_block}

OUTPUT -- only these two XML fields, nothing else:
<reason>concise evidence-grounded rationale with [N] citations, under 120 words</reason>
<answer>A</answer> or <answer>TRUE</answer>
\end{promptbox}

\begin{promptbox}{Interpreter Self-Calibration Prompt Template}
You are the same Interpreter performing a self-calibration step.
The previous answer may be wrong. Re-check the question or claim, answer space,
evidence, and previous answer using ONLY the supplied evidence.

Rules:
- If the previous rationale supports one option but the answer tag names another, correct the answer.
- Match the exact question intent, including why, false, except, not true, deduce, and infer.
- For claim verification, do not add requirements that are not stated in the claim.
- Prefer concrete evidence links over broad or isolated word overlap.
- If another option is better supported by the evidence, change the answer.
- Keep <reason> under 100 words and cite decisive paragraph numbers.

Question or claim: {question}

Answer space:
  (A) {opt_a}
  (B) {opt_b}
  (C) {opt_c}
  (D) {opt_d}
  or TRUE/FALSE for claim verification

Evidence:
{evidence_block}

Previous answer: {previous_answer}
Previous rationale: {previous_reason}

OUTPUT -- only these two XML fields, nothing else:
<reason>self-calibrated rationale with [N] citations</reason>
<answer>A</answer> or <answer>TRUE</answer>
\end{promptbox}

\FloatBarrier

\end{document}